\documentclass{article} 
\usepackage{iclr2026_conference,times}

\usepackage{amsmath,amsfonts,bm}

\def\eqref#1{equation~\ref{#1}}

\def\1{\bm{1}}

\DeclareMathAlphabet{\mathsfit}{\encodingdefault}{\sfdefault}{m}{sl}
\SetMathAlphabet{\mathsfit}{bold}{\encodingdefault}{\sfdefault}{bx}{n}

\usepackage{amsmath}
\usepackage{amssymb}
\usepackage{graphicx}
\usepackage{booktabs}
\usepackage{multirow}
\usepackage{microtype}
\usepackage{hyperref}
\usepackage{url}
\usepackage{subcaption}
\usepackage{xcolor}
\usepackage{wrapfig}

\newcommand{\name}{\textsc{STRAT}}
\newcommand{\names}{\textsc{STRAT's}}

\newcommand{\CE}{\mathrm{CE}}

\title{State Trace Rationale As Auxiliary Task in Reinforcement Learning}

\author{Muhammad U.~Nasir \\
University of the Witwatersrand\\
Johannesburg, South Africa \\
\texttt{muhammad.nasir@wits.ac.za} \\
\And
Alex Vogt\\
University of the Witwatersrand\\
Johannesburg, South Africa \\
\texttt{alex.vogt1@students.wits.ac.za} \\
\And
Steven D.~James\\
University of the Witwatersrand\\
Johannesburg, South Africa \\
\texttt{steven.james@wits.ac.za}\\
\And
Julian Togelius\\
New York University\\
New York, USA \\
\texttt{julian@togelius.com} 
}

\iclrfinalcopy 
\begin{document}

\maketitle

\begin{abstract}
We propose \name{}, an auxiliary task that trains deep reinforcement learning (RL) agents to predict a short textual trace of their own state. 
Inspired by human spatial navigation, the description combines landmark, route, and survey knowledge, tracking the agent's position, inventory, goals, and immediate progress. Environment rules generate this text online without human labelling.
Our method adds a single auxiliary head to a standard policy. 
Across 60 sparse-reward XLand-MiniGrid tasks, \name{} solves complex environments where standard RL fails outright, while compacting state representations and preventing rank collapse. Beyond performance gains, the predicted trace provides a readable account of agent beliefs at every step for no extra cost.

\end{abstract}

\section{Introduction}\label{sec:intro}

Reinforcement learning (RL) algorithms are extremely general, and have been applied to a wide variety of domains such as robotics \citep{kober2013reinforcement}, games \citep{mnih2015human,silver2016mastering}, and language \citep{ouyang2022training,guo2025deepseek}. 
This generality is due to the fact that RL algorithms only require a scalar reward signal to learn from. 
However, when the reward signal is sparse or terminal, it may be difficult or impossible to learn \citep{andrychowicz2017hindsight}. 
In deep RL in particular, optimising for reward means that the learned representation will only capture what predicts return, which may be very little given a sparse signal.

One way to improve learning is to provide auxiliary tasks that leverage the same representation for a second prediction task \citep{jaderberg2016reinforcement}. 
However, deciding what to predict is not straightforward: while early work focused on cheap, task-agnostic targets such as pixels \citep{jaderberg2016reinforcement}, next observations \citep{pathak2017curiosity}, or future latent states \citep{guo2020bootstrap}, recent work has focused on language \citep{lampinen2022tell,mu2022improving}, but has not asked what the description should contain or what parts are important. 

We take inspiration from how people find their way through unfamiliar places. Human spatial navigation is commonly described in terms of three kinds of knowledge~\citep{siegel1975development}: \textit{landmark} knowledge of salient features, \textit{route} knowledge that associates places with the actions taken there, and \textit{survey} knowledge of the overall layout that supports novel paths. 
We propose that an agent should be made to predict exactly these concepts. Our method, \textit{State-Trace Rationale as Auxiliary Task} \name{}, adds a single head to a deep RL agent that predicts a short templated description of the agent's state: its position and heading, the goal, the current subgoal, whether its last action made progress, and what it is carrying.

We make the following contributions.
We introduce (\name{}), which equips agents with an auxiliary head to predict textual state descriptions online directly from environment rules, requiring no human labelling. 
We evaluate our method across 60 \textit{XLand-MiniGrid} tasks \citep{nikulin2024xland} spanning four difficulty tiers, and find that \name{} outperforms a baseline RL algorithm across most of the tasks.
We additionally analyse the learned representations and show that \names{} representations are transparent via probing beliefs, such as if target is visible, making it much more explainable. Because the head is trained to describe the agent's state, our approach also gives a readable account of what the agent believes at every step, at no extra cost.

\section{Background}\label{sec:background}

\subsection{Partially Observable Markov Decision Processes}
A partially observable Markov decision process (POMDP) is a tuple
$\langle \mathcal{S}, \mathcal{A}, \mathcal{O}, T, R, \Omega, \gamma \rangle$.
The environment is in a state $s_t \in \mathcal{S}$ that the agent can partially observe through the observation $o_{t} \sim \Omega(\cdot \mid s_{t})$.
After taking action $a_t$, the state changes to $s_{t+1} \sim T(\cdot \mid s_t, a_t)$,
the agent receives a reward $r_t$ and an observation.
The goal is to maximise the expected discounted return $\mathbb{E}[\sum_t \gamma^t r_t]$.
Because one observation does not fully identify the state, the best action
depends on the whole history $h_t = (o_1, a_1, \ldots, o_t)$, or equivalently
on the belief $b_t = P(s_t \mid h_t)$. Computing the belief exactly is too
expensive in practice, so deep RL methods learn a policy $\pi(a_t \mid h_t)$
that summarises the history in a learned vector. The hidden state, goals and rules require an agent to remember and combine information over time.

\subsection{Memory-Based Policies and Proximal Policy Optimisation}
A memory-based policy uses a sequence model to turn the history $h_t$ into a
vector $z_t$, and then computes the policy $\pi_\theta(a_t \mid z_t)$ and the
value $V_\phi(z_t)$ from it. Recurrent networks~\citep{hausknecht2015deep} are
the classic choice, but they must fit the whole past into one fixed-size
vector and are trained with truncated backpropagation through time.
Transformer-XL~\citep{dai2019transformer} instead keeps a cache of activations from
earlier steps and attends over it, so the model can look back over a long
window at a fixed cost per step. With small changes it trains reliably in
RL~\citep{parisotto2020stabilizing}.
A common approach to training the policy is Proximal Policy
Optimization (PPO)~\citep{schulman2017ppo}, an on-policy actor--critic method.
Advantages $\hat{A}_t$ are computed with generalised advantage
estimation~\citep{schulman2016gae}, and the policy is updated by minimising
the clipped surrogate loss
\begin{equation}
  \mathcal{L}_{\mathrm{PG}}
  = -\,\mathbb{E}_t\!\left[\min\!\left(\rho_t \hat{A}_t,\;
    \operatorname{clip}(\rho_t, 1-\epsilon, 1+\epsilon)\hat{A}_t\right)\right],
  \qquad
  \rho_t = \frac{\pi_\theta(a_t \mid z_t)}{\pi_{\theta_{\mathrm{old}}}(a_t \mid z_t)}.
\end{equation}
The clip stops the policy from changing too much in one update. A value
regression loss $\mathcal{L}_{V} = \mathbb{E}_t\big[(V_\phi(z_t) - \hat{R}_t)^2\big]$,
where $\hat{R}_t = \hat{A}_t + V_{\phi_{\mathrm{old}}}(z_t)$ is the GAE
return, and an entropy bonus
$\mathcal{H} = \mathbb{E}_t\big[\mathcal{H}(\pi_\theta(\cdot \mid z_t))\big]$
are added to this objective, giving
\begin{equation}
  \mathcal{L}_{\mathrm{PPO}}
  = \mathcal{L}_{\mathrm{PG}} + c_v\,\mathcal{L}_{V} - c_e\,\mathcal{H}.
\end{equation}
Because the memory links steps together, each minibatch holds whole
trajectory segments together with the memory they started from, rather than
single shuffled steps.

\subsection{XLand-MiniGrid}
XLand-MiniGrid~\citep{nikulin2024xland} is a set of procedurally
generated gridworld tasks written in JAX. It brings the ideas of
XLand~\citep{team2021open} to MiniGrid~\citep{chevalier2023minigrid}, and since
every environment step is compiled, thousands of environments can run in
parallel on one GPU. Tasks are not written by hand. Instead, each task is a
\emph{ruleset} sampled from a benchmark: a goal predicate from a fixed list
(e.g.\ \texttt{AgentHold}, \texttt{AgentNear}, \texttt{TileNear}) plus a set of
rules that change one tile into another when the agent interacts with it.
Rules can chain, so reaching the goal may need several steps in a fixed
order. The agent is never told the goal or the rules. It sees only a small
symbolic view in front of it and its heading, and receives a single reward at
the end of the episode if the goal is met. It must therefore figure out the
task from the objects it sees and from what happens when it acts on them.. Because the simulator knows
each ruleset's goal and rules, the exact sequence of subgoals needed to solve
it can be computed for every task. We use this to generate the language
targets described above.

\section{Methodology}\label{sec:method}

In this section, we will introduce how \name{} uses rationale as an auxiliary task to improve sample efficiency for PPO. The core idea behind \name{} is to give an agent an understanding of its surroundings with minimal changes. The architectural change is straightforward: \name{} adds one more head that collectively produces probabilities for different aspects of the state. 

\begin{figure}[h]
    \centering
    \includegraphics[width=0.8\textwidth]{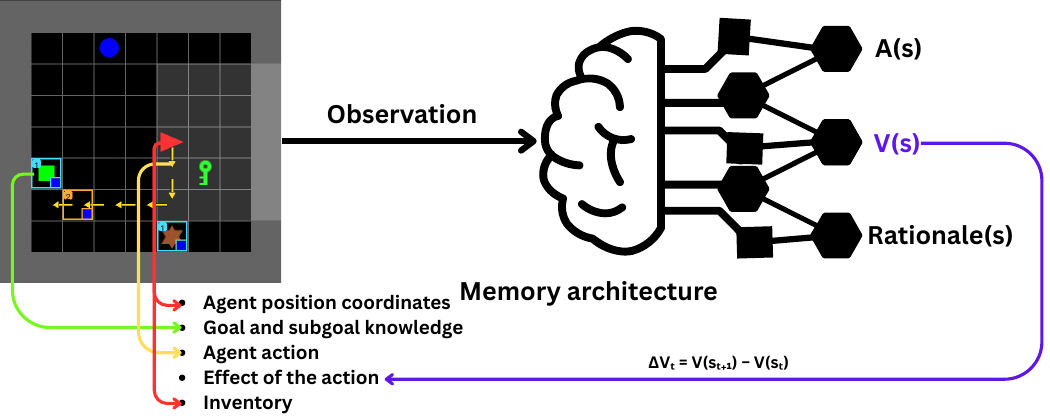}
    \caption{A flow diagram of how \name{} gets the required labels without human intervention.}
    \label{fig:flow}
\end{figure}

\paragraph{Designing labels for state-trace rationale.}

The main contribution of \name{} is the state-trace rationale that is predicted through the \emph{state-trace} auxiliary head. This prediction is a fixed-length sequence of 90 tokens over a grounded vocabulary. Out of the 90 supervised positions, 24 are content slots that vary. The sequence is a templated natural‑language description of the agent's situation composed of five lines, each with a fixed slot structure: an \emph{agent position} line (``I am at position $\langle\text{pos\_y}\rangle$ $\langle\text{pos\_x}\rangle$ facing $\langle\text{direction}\rangle$.''), a \emph{goal} line (``The goal is $\langle\text{goal}\rangle$.''), a \emph{subgoal} line (``The subgoal is to $\langle\text{subgoal}\rangle$.''), an \emph{action} line (``My previous action was $a$. It made me closer/farther/same.''), and an \emph{inventory} line (``Inventory holds $\langle\text{colour}\rangle$ $\langle\text{object}\rangle$.'' or ``Inventory is empty.''). Motivation behind these lines comes from the Human Spatial Navigation framework (see \autoref{sec:intro} for explaination of the framework). Agent position and action line obtains \textit{Route knowledge}, goal and subgoal line gets \textit{Landmark knowledge}, while inventory line adds \textit{Survey knowledge} to the representation. Only the content slots vary; template tokens are constant. Labels are generated online during the rollout from the simulator state and the task specification, and are never shown to the policy as input. The agent line is filled from the agent's global position and heading in the environment state.
The goal line is derived from the sampled ruleset's goal encoding, which we render into tokens (e.g.\ ``near purple square''). The subgoal line is obtained from a causal plan precomputed for each ruleset by backward-chaining from the goal through the ruleset's production rules; the plan is a sequence of phases, each associated with a target tile, and the current phase is tracked online by detecting when the corresponding production has fired in the grid (or the
required object has entered the inventory). The action slot is the executed action; the progress slot is labelled from the critic's own value estimates as the sign of $\Delta V_t = V(s_{t+1}) - V(s_t)$ thresholded at $\tau$, so that ``closer'', ``farther'', and ``same'' reflect learned value progress rather than a geometric distance. The inventory line is read from the agent's pocket after the action.

\paragraph{Loss Function.}

For the loss, let $\mathcal{P}$ denote the set of supervised state-trace token positions, $L = |\mathcal{P}|$, and $\mathcal{V}$ the state-trace vocabulary. At timestep $t$,
the state-trace head maps the memory architecture's hidden state $h_t$ to logits
$z_{t,\ell} \in \mathbb{R}^{|\mathcal{V}|}$ for each $\ell \in \mathcal{P}$,
and $y_{t,\ell} \in \mathcal{V}$ is the target token derived from the
environment state. The trace loss is the token-level cross-entropy between
the predicted distribution and the target, averaged over positions:
\begin{equation}
  \CE_{t,\ell}
  = -\log p_\theta\!\left(y_{t,\ell} \mid h_t\right),
  \qquad
  p_\theta(v \mid h_t) = \mathrm{softmax}(z_{t,\ell})[v],
\end{equation}
\begin{equation}\label{eq:4}
  \mathcal{L}_{\mathrm{ST}}(\theta)
  = \mathbb{E}_{t}\!\left[
      \frac{1}{L} \sum_{\ell \in \mathcal{P}} \CE_{t,\ell}
    \right],
\end{equation}
where the expectation is over active timesteps of on-policy rollouts. The full
objective becomes:
\begin{equation}
  \mathcal{L}
  = \mathcal{L}_{\mathrm{PG}} + c_v\,\mathcal{L}_{V} - c_e\,\mathcal{H}
  + \lambda_{\mathrm{st}}\,\mathcal{L}_{\mathrm{ST}}.
\end{equation}

\section{Experimentation}
\label{sec:experiments}

In this section, we will describe our experimental setup. Our experiments addresses the important question: does \name{} improve sample efficiency compared to an identical PPO agent? Moreover, we perform extensive ablation studies to observe what aspect of \name{} matters the most. 

\subsection{Environment and tasks}
All experiments use XLand-MiniGrid~\citep{nikulin2024xland}, a JAX-native
suite of procedurally generated grid-world tasks. Each task is defined by a
\emph{ruleset}: a goal predicate (e.g.\ \emph{AgentNear}, \emph{TileNear},
\emph{AgentHold}) together with a set of production rules that transform
objects when they are brought into contact. Rules are hidden from the agent,
so solving a task requires discovering the causal chain that produces the goal
object. We use the $9{\times}9$ environment
(\texttt{XLand-MiniGrid-R1-9x9}) with the symbolic observation, a $5{\times}5$
egocentric window of (tile, colour) pairs plus the agent's heading, an action
space of six primitives (move forward, turn left/right, pick up, put down,
toggle), and a per-episode horizon of $243$ steps. Reward is sparse: a
single terminal reward of $1 - 0.9\,t/T_{\max}$ on success and $0$ otherwise. We select 20 environments each from \emph{small-1m}, \emph{medium-1m}, and \emph{high-1m} benchmarks. We sample five tasks from each family in \autoref{tab:task-fam} on each of the three benchmarks: 15 tasks per family and 60 in total.

\begin{table}[t]
    \caption{Ruleset structure families. A task takes the first matching row. Placed is decided by whether any rule produces a goal tile; the other three are read from the backward chain, whose last phase is the goal.}
    \label{tab:task-fam}
    \centering
    \small
    \begin{tabular}{llp{0.58\linewidth}}
        \toprule
        Family & Difficulty & Definition \\
        \midrule
        Placed & Easy & No rule produces the goal tile, so it is already on the grid. The goal is reach or hold. \\
        Go-hold & Medium & The goal tile is crafted. The goal is reach or hold, and every earlier step is only walk or pick up. \\
        Beside & Hard & The goal tile is crafted. The goal is ``put $A$ beside $B$,'' and every earlier step is only walk or pick up. \\
        Align & Hardest & The goal tile is crafted, and some earlier step must line two objects up to produce a tile. \\
        \bottomrule
    \end{tabular}
\end{table}

\subsection{Agent and optimisation}
\label{sec:exp-agent}
The agent is a Transformer-XL~\citep{dai2019transformer} actor-critic. The symbolic
observation is encoded by an entity/colour embedding followed by a small
convolutional stack; the resulting features are concatenated with embeddings
of the agent's heading, the previous action, the previous reward, and a
two-dimensional episode descriptor, i.e episode progress, and projected
to the transformer width. The TXL memory cache is sized to cover the full
episode, so the policy can attend to every step it has taken since the start
of the episode. Policy and value heads are single-hidden-layer MLPs on
the final transformer features. The state-trace head applies a hidden
projection to the same features, adds a learned embedding for each supervised
trace position, and maps each position through a shared linear layer onto the
token vocabulary (\autoref{sec:method}).

All conditions are trained with PPO and generalised
advantage estimation, the only difference between
conditions is the auxiliary term of the loss. Rollouts are collected from parallel environments on a single GPU, the learning rate is decayed
linearly to zero, and each configuration is repeated over 10 random seeds.
All curves and tables report the mean and interquartile range across seeds. Architecture, optimisation, and environment hyperparameters are
listed in full in \autoref{app:hparams} (\autoref{tab:hparams}).

\begin{figure}[h]
    \centering
    \includegraphics[width=1\textwidth, height=0.46\textheight]{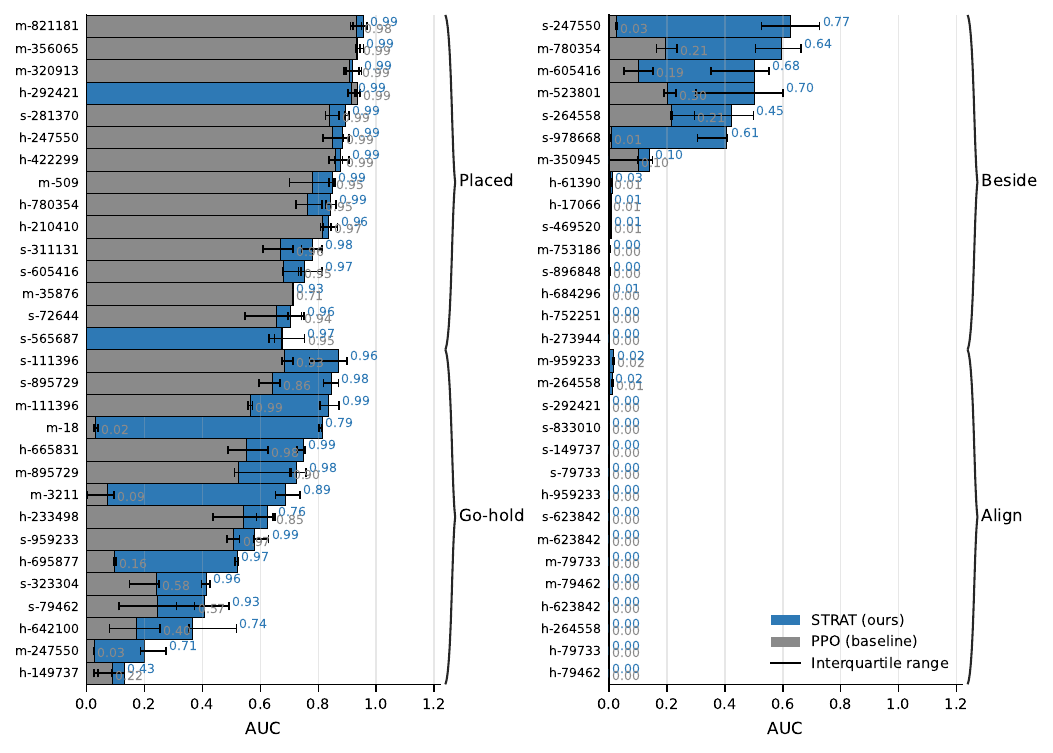}
    \caption{Comparison of \name{} and PPO across 60 tasks. The bars represents mean normalized AUC of the learning curve. The whiskers are the interquartile range for 10 seeds. The numbers represent the mean final returns.}
    \label{fig:results_bar}
\end{figure}

\subsection{Ablations}\label{sec:exp-ablations}

We perform the following ablations:

\begin{enumerate}
    \item \textbf{Different combinations of state-trace rationales} to mimic the Human Spatial Navigation (HSN) framework (see \autoref{sec:intro} for the definition of different types of knowledge), and the impact of each knowledge category:
    \begin{enumerate}
        \item \textit{Goal and sub-goal only:} Here we want to see whether goal and sub-goal provide \name{} with enough rationale to beat the baseline. This is akin to the agent only having \textit{Landmark Knowledge}.
        \item \textit{\names{} position and action description only:} Here we want to see if \name{} can still learn without goal, sub-goal, and inventory rationales. Position and action knowledge can be considered as \textit{Route Knowledge}, as it can give the agent a place association by knowing the coordinates and if the actions are making the agent closer or not.
        \item \textit{Without \names{} action description:} Here we want to see if the value-bootstrapped action rationale provides a good signal for \name{}. These auxiliary tasks give the agent the \textit{Landmark}, parts of \textit{Route} and \textit{Survey} knowledge. Survey Knowledge comes from knowing inventory and subgoals, as these can allow the agent to know about the features of the environment.
        \item \textit{Without \names{} position description:} Here we want to see if the position knowledge provides a good signal for \name{}. We want to see if having less Route Knowledge via having less position awareness will impact the results.
    \end{enumerate}
    \item \textbf{Action rationale} by replacing value-bootstrapped rationale with Breadth-First Search (BFS) and Euclidean distance. This is an important ablation as some RL algorithms are critic-free, and if we know the geometry of the environment, we can use a distance metric or a search algorithm. 
    \item \textbf{RNN memory backbone} to demonstrate that \name{} works with any type of memory architecture, we will use Gated Recurrent Unit (GRU)~\citep{cho2014properties}, a type of Recurrent Neural Network (RNN)~\citep{chung2014empirical}.
\end{enumerate}

\subsection{Representation analysis}\label{sec:repr-analysis}
We analyse frozen checkpoints saved at 40 evenly spaced points during training, for \name{},PPO, and ablations for \name{}, 10 seeds each. Each checkpoint is run with the segment length and attention-memory length used in training.

We read two representations from each checkpoint. The first is the trunk output: the vector the actor, the critic and the state-trace head all consume. The second is the per-step observation encoding that enters the memory backbone, which has no temporal context. Comparing the two tells us whether the recurrent trunk creates a difference between \name{} and the PPO baseline or whether it is already present in the per-step encoding.
\subsubsection{Minimum Description Length (MDL)}
Accuracy cannot tell apart a variable that is easy to extract from one that a probe recovers only with a lot of data. We therefore also report the prequential (online) codelength of the labels given the features~\citep{voita2020information}. The probe is refitted on growing prefixes of the data, each block doubling in size from 0.2\% of the data. It is charged the codelength of each next block, and we report compression relative to a uniform code. Higher compression means the variable is more readily available in the representation.
\subsubsection{Representational Capacity}
To test whether the auxiliary objective prevents representational collapse, we track the dimensionality of the trunk over training with four statistics:
\begin{itemize}
    \item {\bf{Effective rank}}: the exponential of the entropy of the normalized singular-value spectrum~\citep{roy2007effective}
    \item {\bf{srank}}: the number of singular values needed to capture 99\% of the spectrum  
    \item {\bf{participation ratio}} of the covariance eigenvalues
    \item {\bf{dormant units}}: the fraction of units whose mean activation falls below 2.5\% of the layer average~\citep{sokar2023dormant}
\end{itemize}
All four are also computed on the per-step encoding, so that any collapse can be located in the trunk or the encoder. Capacity also rises when a run learns to complete the task. 





\section{Results}

\subsection{\name{} vs PPO}

We start this section by comparing \name{} with a similar PPO agent as a baseline. The \autoref{fig:results_bar} shows performance on 60 tasks. The bars show the AUC of the learning curves, and the whiskers show the interquartile range. The numbers show the achieved final return. We clearly observe a pattern: \textit{easy} tasks from the \textit{Placed} family across all benchmarks are easy for both methods to solve. The reason is that they are straightforward, as the goal is already placed. \textit{Go-hold} is where the major difference lies. The performance of \name{} is strictly better on all of the tasks and is significantly better on many of them. These tasks are not straightforward to solve. You either have to reach a tile that may or hold it, but the agent may need to hold another tile to fulfil the goal, which makes it nontrivial to solve. The tasks that fall under \textit{Beside} classification are \textit{hard} tasks, and both of the methods failed to achieve good results, yet \name{} achieves considerably better returns compared to PPO in almost half of the environments. \textit{Align} family of tasks are the \textit{hardest}, and we see a collapse of performance for both methods. The reason is that there are no rewards for aligning two tiles as sub-goals, and the agent still needs to craft the goal tile. Previous works have shown that these kinds of tasks are learnt via curriculum learning or open-ended learning~\citep{wang2019paired, team2021open, team2023human}.

\begin{figure}[t]
    \centering
    \begin{subfigure}[t]{0.24\textwidth}
        \centering
        \includegraphics[width=\linewidth]{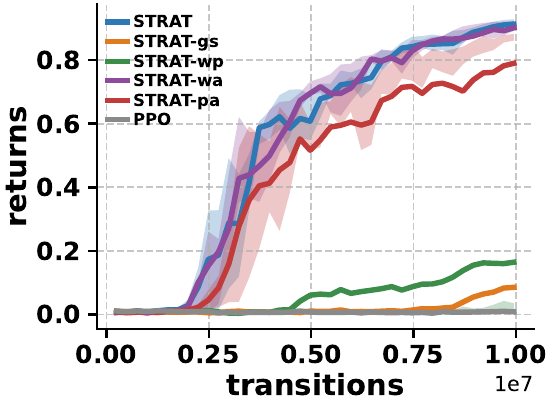}
        \caption{\name{} ablation on m-3211 (Go-hold). Mean with interquartile ranges as the shaded region.}
        \label{fig:results_sab1}
    \end{subfigure}\hfill
    \begin{subfigure}[t]{0.24\textwidth}
        \centering
        \includegraphics[width=\linewidth]{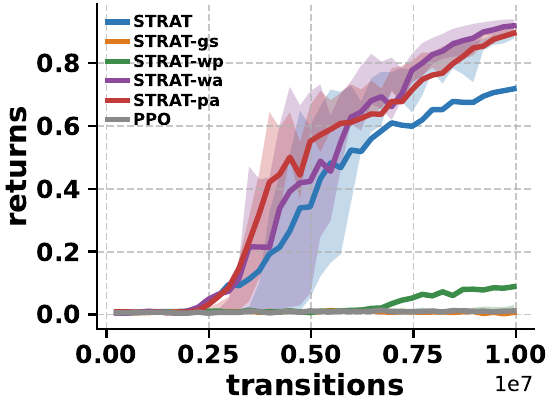}
        \caption{\name{} ablation on m-18 (Go-hold).}
        \label{fig:results_sab2}
    \end{subfigure}\hfill
    \begin{subfigure}[t]{0.24\textwidth}
        \centering
        \includegraphics[width=\linewidth]{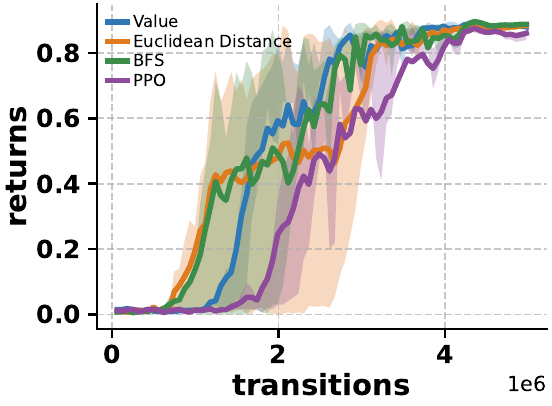}
        \caption{Value bootstrap ablation on s-1134 (Go-hold).}
        \label{fig:results_value_1}
    \end{subfigure}\hfill
    \begin{subfigure}[t]{0.24\textwidth}
        \centering
        \includegraphics[width=\linewidth]{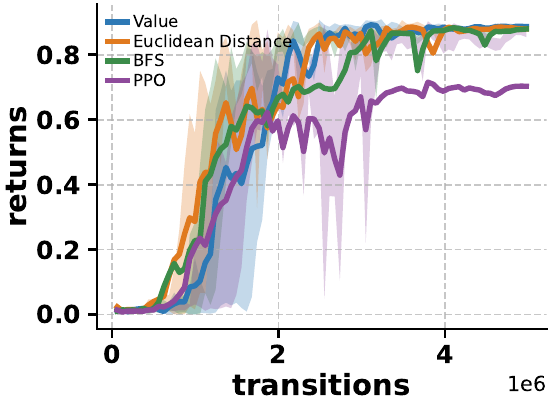}
        \caption{Value bootstrap ablation o s-999911 (Beside).}
        \label{fig:results_value_2}
    \end{subfigure}\hfill

    \vspace{0.6em}
    \begin{subfigure}[t]{0.24\textwidth}
        \centering
        \includegraphics[width=\linewidth]{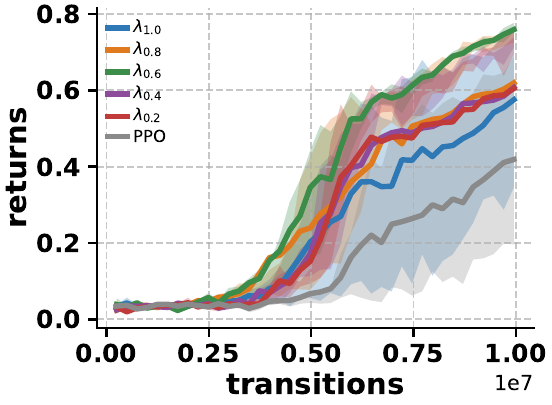}
        \caption{Effect of $\lambda_{st}$ on h-642100 (Go-hold).}
        \label{fig:results_lambda_1}
    \end{subfigure}\hfill
    \begin{subfigure}[t]{0.24\textwidth}
        \centering
        \includegraphics[width=\linewidth]{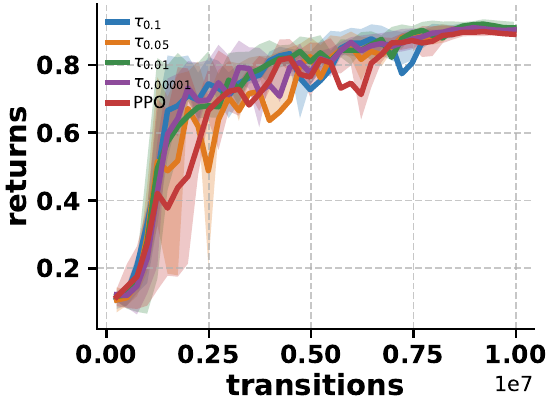}
        \caption{Effect of $\tau$ on s-117 (Placed).}
        \label{fig:results_tau_1}
    \end{subfigure}
    \begin{subfigure}[t]{0.24\textwidth}
        \centering
        \includegraphics[width=\linewidth]{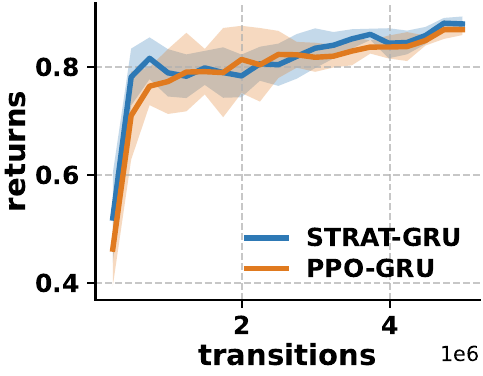}
        \caption{GRU on s-281370 (Placed). \name{}-TXL achieved a return of $0.99$ here.}
        \label{fig:results_rnn_1}
    \end{subfigure}\hfill
    \begin{subfigure}[t]{0.24\textwidth}
        \centering
        \includegraphics[width=\linewidth]{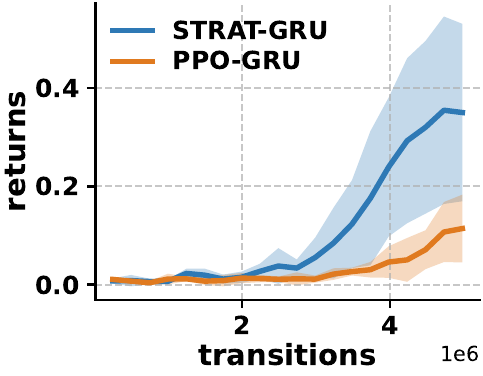}
        \caption{GRU on s-264558 (Beside). \name{}-TXL achieved a return of $0.77$ here.}
        \label{fig:results_rnn_2}
    \end{subfigure}
    \caption{Ablations, $\lambda_{st}$ and $\tau$ sweeps, and GRU learning curves.}
    \label{fig:results_sweeps}
\end{figure}

\subsection{Ablations}

\paragraph{State-trace rationales.} This is the most important ablation for \name{}.  \autoref{fig:results_sab1} shows results on m-3211, a task from \textit{Go-hold} family which is a hard task. Here, \name{}-gs means \textit{goal and sub-goal only}, \name{}-pa means \textit{\names{} position and action description only}, \name{}-wa means \textit{without \names{} action description}, and \name{}-wgs means \textit{\name{} without goals and subgoals}. Notably, the baseline performance collapses. \name{}-gs and \name{}-wp also performs significantly worse. \name{} and the versions without action description and with only position and action are performing well. Similar results are observed for another \textit{Go-hold} task (m-18), but this time PPO and \name{}-gs collapse, while \name{}-wp is worse than m-3211. Other results are similar yet \name{} performs worse than other two versions. $25^{th}$ and $75^{th}$ percentile sits with other versions but mean is lower.

The important observation here is that providing learning signals such as goals and subgoals alone does not help. In contrast with Human Spatial Navigation (HSN) abilities, these are called \textit{landmark Knowledge} (see \autoref{sec:intro}). This is intuitive: the results indicate that this information will eventually lead the agent to the goal, but it will take much more time than otherwise, which is why we get a mean return of around 0.1. Another intuitive result is that you the agent needs to have position knowledge. This is \textit{Route Knowledge} in the HSN framework. If the agent can not associate where it is standing, then the performance decreases. \name{}-pa (position and action knowledge only) and \name{}-wa (without action knowledge) are also intuitive. With position and action knowledge. you have complete Route knowledge; therefore, you should be able to reach your goal, and with the help of PPO backbone algorithm, you can perform tasks. While \name{}-wa, which has position, goal and subgoal, and inventory knowledge, means that the agent has full landmark knowledge, some route and some survey knowledge (via what inventory holds). This kind of knowledge should able human to traverse through unknown places. This result shows us why all the aspects of auxiliary tasks in \name{} are needed to solve these complex tasks efficiently.

\paragraph{RNN backbone.} As the focus of our methodology is towards auxiliary head, it makes sense to ablate the backbone memory architecture. As mentioned in \autoref{sec:exp-ablations}, we are using GRU as the memory architecture. \autoref{fig:results_rnn_1} shows results on an easy task (s-281370, Placed), while  \autoref{fig:results_rnn_2} shows performance of GRU on s-264558, a substantially difficult task from \textit{Beside} family of task. This demonstrates that any memory backbone can be used.   

\paragraph{Action rationale and effects of $\lambda_{ST}$ and $\tau$.} Here we replace value bootstrap with Breadth-First Search (BFS) and Euclidean distance. \autoref{fig:results_value_1} and \ref{fig:results_value_2} shows that there is not much difference in performance in BFS and euclidean distance. In some cases it proves to be better than value bootstrapping. $\lambda_{ST}$ is the weight we give to the state-trace loss, and $\tau$ is the threshold for $\Delta V$. We observe that the results are similar for sample efficiency but different in the final return achieved, as suggested by \autoref{fig:results_lambda_1}, and \ref{fig:results_tau_1}. These should be considered as hyperparameters that need to be tuned

\begin{figure}[t]
    \centering
    \begin{subfigure}[t]{0.44\textwidth}
        \centering
        \includegraphics[width=\linewidth]{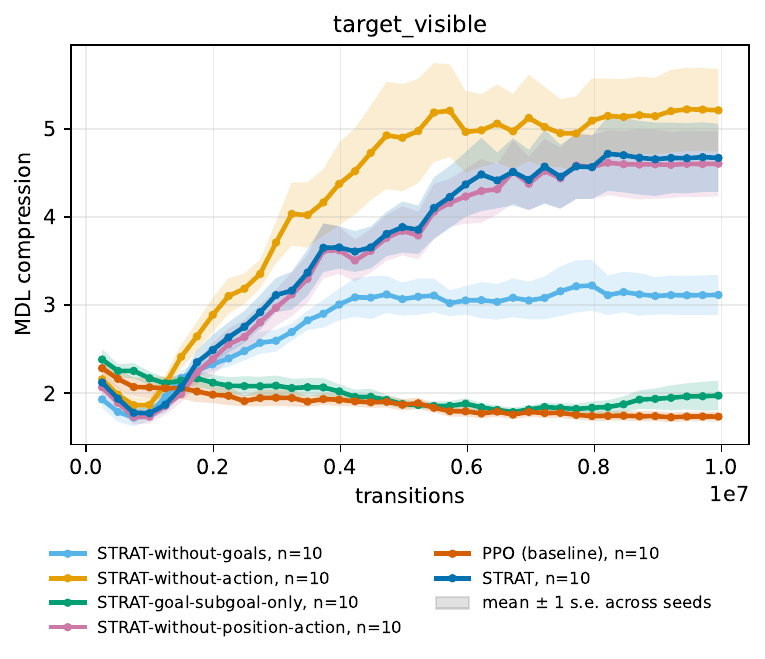}
        \caption{Target visibility compressed in representation.}
        \label{fig:results_mdl_target}
    \end{subfigure}
    \begin{subfigure}[t]{0.44\textwidth}
        \centering
        \includegraphics[width=\linewidth]{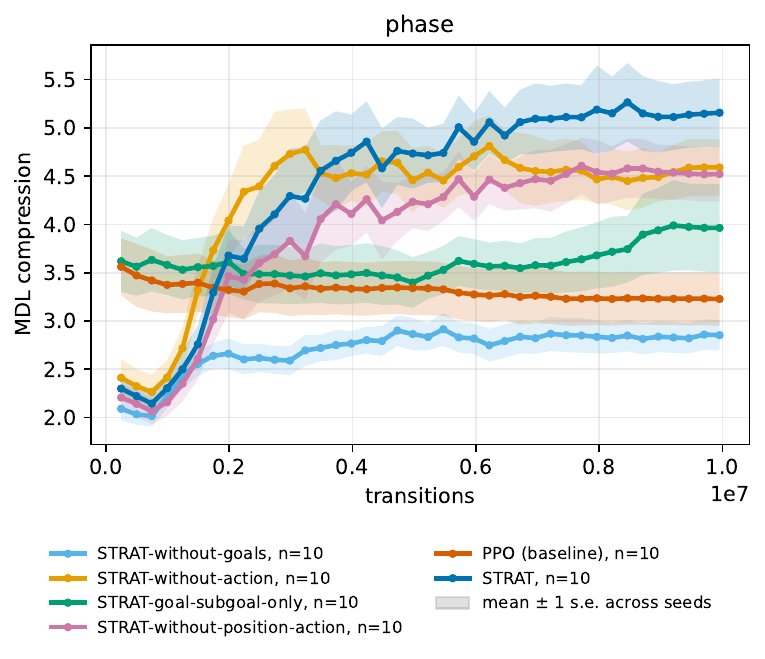}
        \caption{Subgoal phase compressed in representation.}
        \label{fig:results_mdl_phase}
    \end{subfigure}
    \caption{Information compressed in the representation.}
    \label{fig:results_mdl}
\end{figure}

\subsection{Learnt Representations}

\begin{figure}[t]
    \centering
    \begin{subfigure}[t]{0.45\textwidth}
        \centering
        \includegraphics[width=\linewidth]{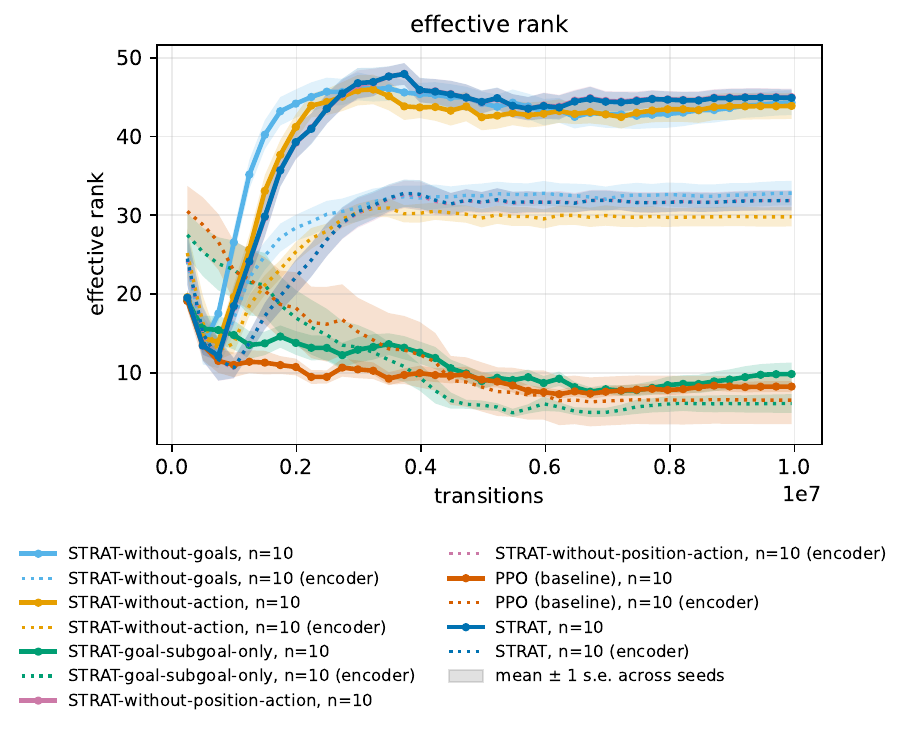}
        \caption{Effective rank.}
        \label{fig:results_capacity_erank}
    \end{subfigure}
    \begin{subfigure}[t]{0.45\textwidth}
        \centering
        \includegraphics[width=\linewidth]{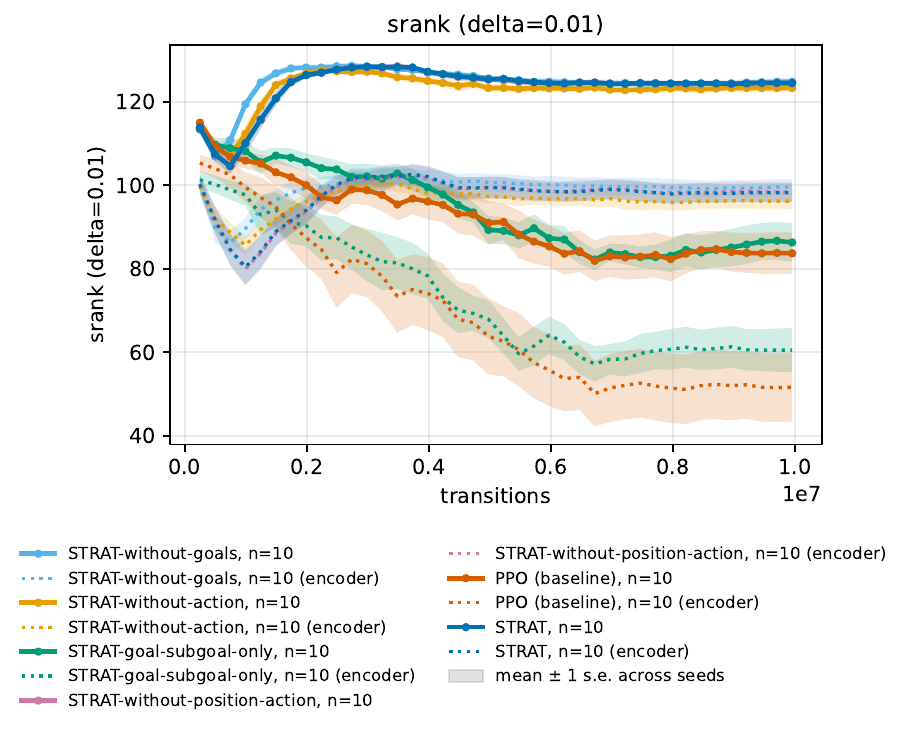}
        \caption{Srank.}
        \label{fig:results_capacity_srank}
    \end{subfigure}
    \caption{Information compressed in the representation capacity.}
    \label{fig:results_capacity_1}
\end{figure}

As described in \autoref{sec:repr-analysis}, Minimum Description Length (MDL) is the measure of how much a variable is present in the representation. \autoref{fig:results_mdl_target} shows target visibility in the representations. We observe that \name{} (dark blue), \name{}-wa (yellow), and \name{}-wp (pink) steadily compress this information, while \name{}-wg (light blue) plateaus after a certain degree of compression. \name{}-gs (green) and PPO fail to compress this information. This means that goal and subgoal alone are not what is needed for the representation to learn target visibility but other rationales also matter. This result indicates that to get to the \textit{landmark} agent has to have either all spatial knowledge (\name{}) or position knowledge and goal knowledge (\name{}-wa), or if the agent knows that it is getting closer to the target (\name{}-wp). This is intuitive if we see it through the lens of the Human Spatial Navigation (HSN) framework. If a human has the \textit{landmark knowledge} but no \textit{route knowledge}, then finding the landmark in an unknown place will become difficult. Humans need some way to know that they are getting closer to the goal, either by knowing that the distance is getting smaller (\textit{action rationale} in \name{}) or by knowing their own place (\textit{position rationale} in \name{}). Similarly, phase of the subgoal is also compressed smoothly in \name{}, as shown in \autoref{fig:results_mdl_phase}. \name{}, having all rationales, compresses phase information well, while not having goal information makes \name{} worse than PPO. Having only goal and subgoal rationale is also not enough. Notably, PPO does not compress phase information at all. In \autoref{app:repr} we have more compression analysis for more variables, such as action, agent direction, agent position, and inventory information. With all of them, we see a similar pattern where \names{} observed representation creates a transparent belief of the variable in focus; therefore, it is a representation that is more explainable compared to PPO, where explainability is much less. \autoref{fig:results_capacity_1} shows the representational capacity. Here, we can observe that effective rank and srank (see ~\autoref{sec:method} for definitions) do not collapse for \name{}, and have a major representation difference from PPO. \autoref{fig:results_capacity_1} in \autoref{app:repr} shows that the participation ratio is similar to effective and srank, while in dormant units, we see that there are no units for \name{} where the mean activation falls below $2.5\%$.

\section{Related Works}

Our work is closely related to auxiliary tasks and language rationales in Reinforcement Learning. In this section, we will look into the works that fall into these categories and are related to our work.


\subsection{Auxiliary Tasks in Reinforcement Learning}

When reward is sparse, the agent gets little feedback and the learned
representation tends to capture only what predicts return. Auxiliary tasks
add extra prediction objectives that share the agent's representation and are
trained together with the RL loss,
$\mathcal{L} = \mathcal{L}_{\mathrm{RL}} + \beta\, \mathcal{L}_{\mathrm{aux}}$.
The auxiliary heads are only used during training and do not affect how the
agent acts at inference time. Early examples predicted pixel changes and
reward~\citep{jaderberg2016reinforcement}, depth~\citep{mirowski2017learning}, or the
next state and the action taken~\citep{pathak2017curiosity}; later work
predicted the model's own future
representation~\citep{schwarzer2020data} or used contrastive
objectives~\citep{laskin2020curl}. These tasks help most when the target
matters for solving the task and cannot be read directly off the current
observation, so predicting it forces the agent to remember the past or infer
what is hidden. Recent work uses natural language as the
target~\citep{lampinen2022tell}. A sentence can state what the
task is, what to do next, and whether the last action helped, and such
sentences can be generated automatically from the simulator's internal state
without human labelling. This is the approach we take.

\paragraph{Language rationale in RL}

When the extrinsic reward does not identify which part of a long behaviour matters, the goal, the subgoal, or the reason for an action can give a policy compositional intermediate rewards. \citet{andreas2017modular} made this claim with policy sketches: each task is annotated with a sequence of named subtasks, and each name is bound to a modular sub-policy that is shared across tasks. \citet{andreas2018learning} showed that the description itself can be the object of learning. A model is pretrained to interpret language, and a new behaviour is then obtained by searching for a string that minimises the interpreter's loss, so language is required only in that pretraining stage. \citet{jiang2019language} placed language between a high-level policy and an instruction-following low-level policy, and found that the compositional form of the instruction is what lets sub-skills recombine on new tasks. The same grounding problem, in grid worlds with a fragment of English, is the subject of BabyAI~\citep{chevalierboisvert2019babyai}. Later work generates the rationale in the loop with acting. ReAct interleaves free-form reasoning traces with environment actions~\citep{yao2023react}, and Reflexion writes a verbal critique of a failed episode and feeds that text back as the context of the next attempt~\citep{shinn2023reflexion}.

\section{Conclusion}

To conclude, we present \name{}, an agent that adds an auxiliary head that include state-trace rationale in the common learned representation. We take this inspiration from Human Spatial Navigation~\cite{siegel1975development}. We experiment on 60 tasks from three XLand-minigrid benchmarks as they have long-horizon tasks with extremely delayed and sparse rewards. We further categorise these tasks into 4 tiers of difficulty. We observe that \name{} outperforms PPO in $100\%$ of the medium-difficulty (\textit{Go-hold}) tasks, while doing better on $40\%$ of the hard tasks ($Beside$), where the rest are unsolvable for both. Whereas hardest ($Align$) difficulty remains unsolvable for both of the methods and easy ($Placed$) tasks equally solved by both of the methods. We also perform extensive ablations to study the impact of each rationale. Furthermore, we observe the learnt representation through minimum description length and representational capacity. We found that \name{} representation is transparent towards many different aspects of the state, such as target visibility, phase of the subgoals etc. We conclude that the representation is much more explainable than PPO.

\section{AI use Statement}

In this work, we used generative AI tools for coding help. Code is not written completely by AI. All LLM-generated code has been carefully reviewed. It is just used to debug the code. We have not used generative AI tools for any writing or ideation purposes.

\bibliography{iclr2026_conference}
\bibliographystyle{iclr2026_conference}

\newpage
\appendix
\section{Hyperparameters}\label{app:hparams}
You may include other additional sections here.

\begin{table}[h]
\centering
\small
\setlength{\tabcolsep}{6pt}
\begin{tabular}{@{}lcc@{}}
\toprule
 & \texttt{small-1m} & \texttt{medium-1m} / \texttt{high-1m} \\
\midrule
\multicolumn{3}{@{}l}{\textit{Environment}} \\
Environment                     & \multicolumn{2}{c}{\texttt{XLand-MiniGrid-R1-9x9}} \\
Observation                     & \multicolumn{2}{c}{symbolic $5{\times}5$ egocentric (tile, colour) + heading} \\
Actions                         & \multicolumn{2}{c}{6} \\
Episode length       & \multicolumn{2}{c}{243} \\
Reward                          & \multicolumn{2}{c}{$1 - 0.9\,t/T_{\max}$ on success, else 0} \\
\midrule
\multicolumn{3}{@{}l}{\textit{Architecture}} \\
Obs.\ embedding dim (entity, colour) & 16 & 16 \\                                  
Obs.\ conv.\ stack              & \multicolumn{2}{c}{$2{\times}2$ conv, 16--32--64 ch., ReLU, VALID} \\
Action embedding dim            & 16 & 16 \\                                       
TXL layers                      & 1 & 3 \\                                         
TXL hidden width $D$            & 72 & 144 \\                                     
TXL attention heads             & 2 & 6 \\                                         
TXL feed-forward width          & 144 & 864 \\                                     
TXL memory cache (per layer)    & \multicolumn{2}{c}{81} \\
Positional encoding             & \multicolumn{2}{c}{relative (Transformer-XL)} \\
Policy / value head hidden width & 128 & 512 \\                                    
Head activation                 & \multicolumn{2}{c}{GELU} \\
State-trace head hidden width   & 128 & 512 \\                                     
State-trace positions  & \multicolumn{2}{c}{90} \\
Vocabulary (word + number tokens) & \multicolumn{2}{c}{$86 + 16 = 102$} \\        
\midrule
\multicolumn{3}{@{}l}{\textit{PPO optimisation}} \\
Parallel environments           & 256 & 1024 \\
Steps per update (per env)      & \multicolumn{2}{c}{81} \\
Inner updates per episode & \multicolumn{2}{c}{3 / 19} \\
PPO epochs                      & \multicolumn{2}{c}{1} \\
Minibatches                     & \multicolumn{2}{c}{16} \\
Minibatch size (envs)           & 16 & 64 \\
Clip ratio $\epsilon$           & \multicolumn{2}{c}{0.2} \\
Discount $\gamma$               & \multicolumn{2}{c}{0.999} \\
GAE $\lambda$                   & \multicolumn{2}{c}{0.95} \\
Value loss coefficient          & \multicolumn{2}{c}{0.5} \\
Entropy coefficient             & \multicolumn{2}{c}{0.05} \\
Auxiliary weight $\lambda_{\mathrm{st}}$ & \multicolumn{2}{c}{1.0} \\
Optimiser                       & \multicolumn{2}{c}{Adam ($\varepsilon = 10^{-8}$)} \\
Learning rate                   & \multicolumn{2}{c}{$3\times10^{-3}$, linear decay to 0} \\
Max.\ gradient norm             & \multicolumn{2}{c}{0.5} \\
Total environment steps         & \multicolumn{2}{c}{$10^7$} \\
Seeds                           & \multicolumn{2}{c}{$\{1, 7, 11, 32, 42, 47, 101, 123, 127, 145\}$} \\
\midrule
\multicolumn{3}{@{}l}{\textit{State-trace labels}} \\
Progress threshold $\tau$       & \multicolumn{2}{c}{$10^{-5}$} \\
\bottomrule
\end{tabular}
\caption{Full hyperparameter settings. Values spanning both columns are shared
by all benchmarks.}
\label{tab:hparams}
\end{table}

\newpage
\section{learnt representations}\label{app:repr}

\begin{figure}[h]
    \centering
    \begin{subfigure}[t]{0.48\textwidth}
        \centering
        \includegraphics[width=\linewidth]{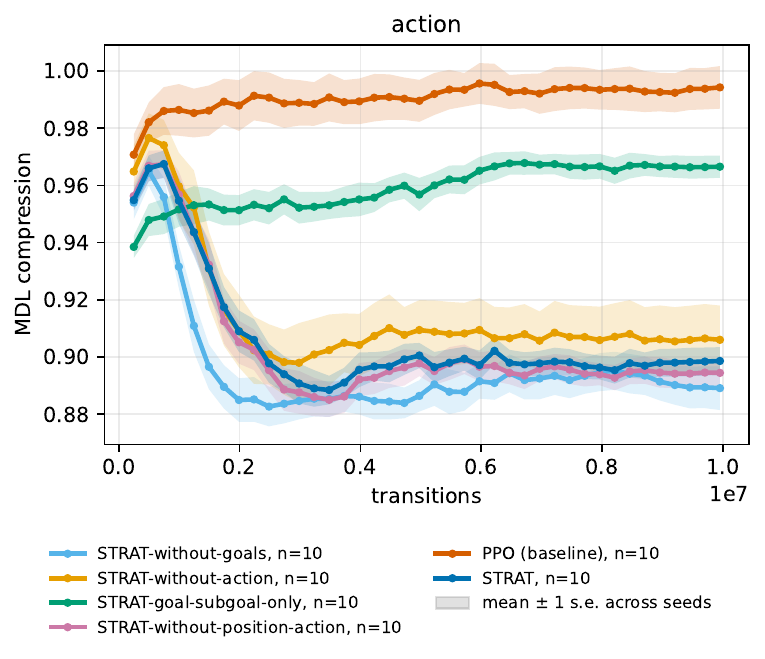}
        \caption{Action compressed in the representation.}
        \label{fig:results_mdl_action}
    \end{subfigure}\hfill
    \begin{subfigure}[t]{0.48\textwidth}
        \centering
        \includegraphics[width=\linewidth]{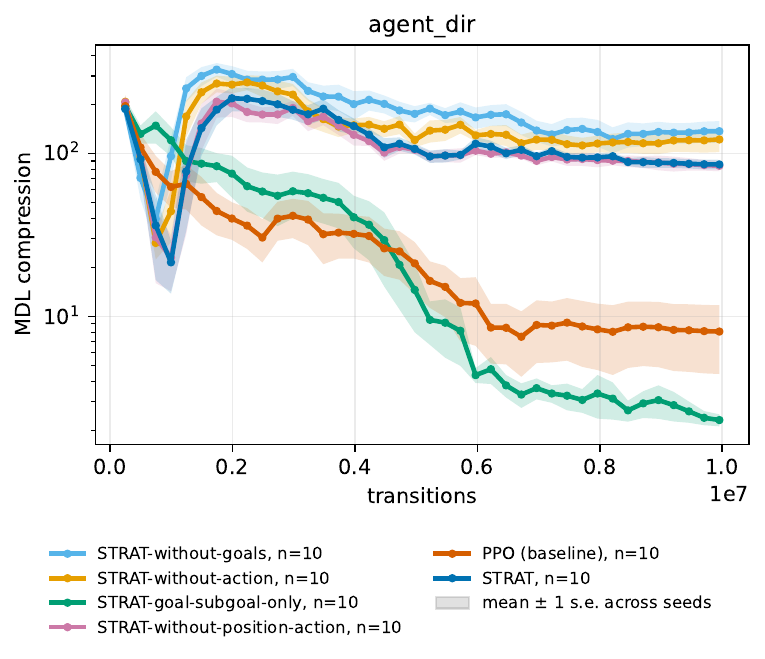}
        \caption{Agent direction compressed in the representation.}
        \label{fig:results_mdl_agent_dir}
    \end{subfigure}

    \vspace{0.6em}
    \begin{subfigure}[t]{0.48\textwidth}
        \centering
        \includegraphics[width=\linewidth]{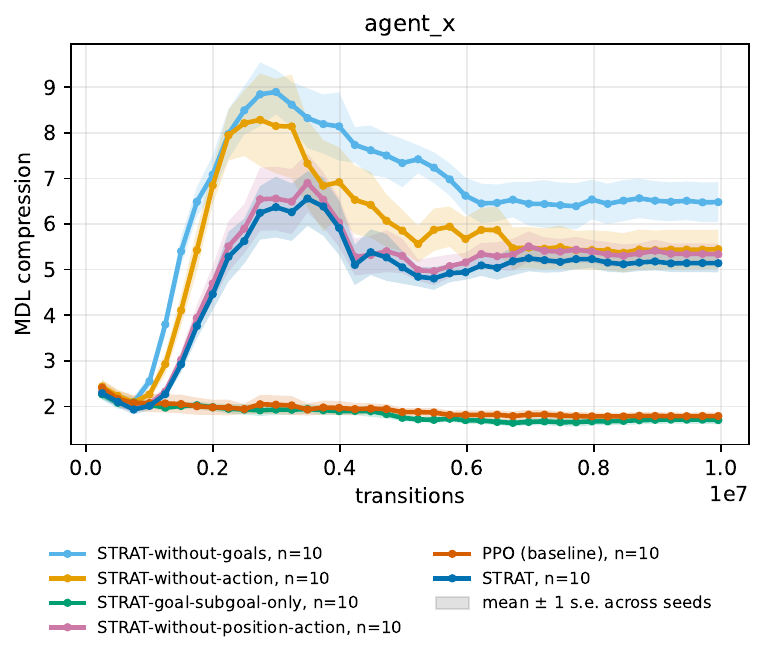}
        \caption{Agent $x$ compressed in the representation.}
        \label{fig:results_mdl_agent_x}
    \end{subfigure}\hfill
    \begin{subfigure}[t]{0.48\textwidth}
        \centering
        \includegraphics[width=\linewidth]{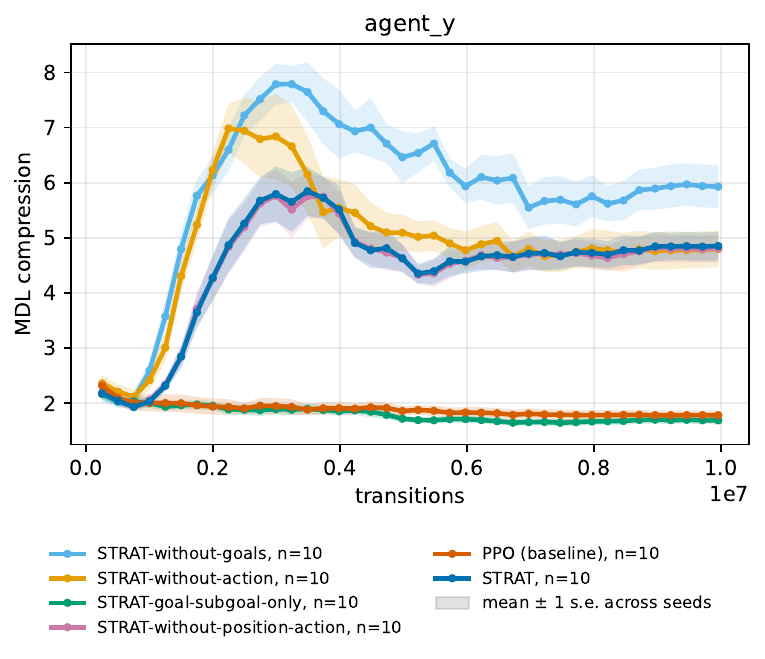}
        \caption{Agent $y$ compressed in the representation.}
        \label{fig:results_mdl_agent_y}
    \end{subfigure}

    \vspace{0.6em}
    \begin{subfigure}[t]{0.48\textwidth}
        \centering
        \includegraphics[width=\linewidth]{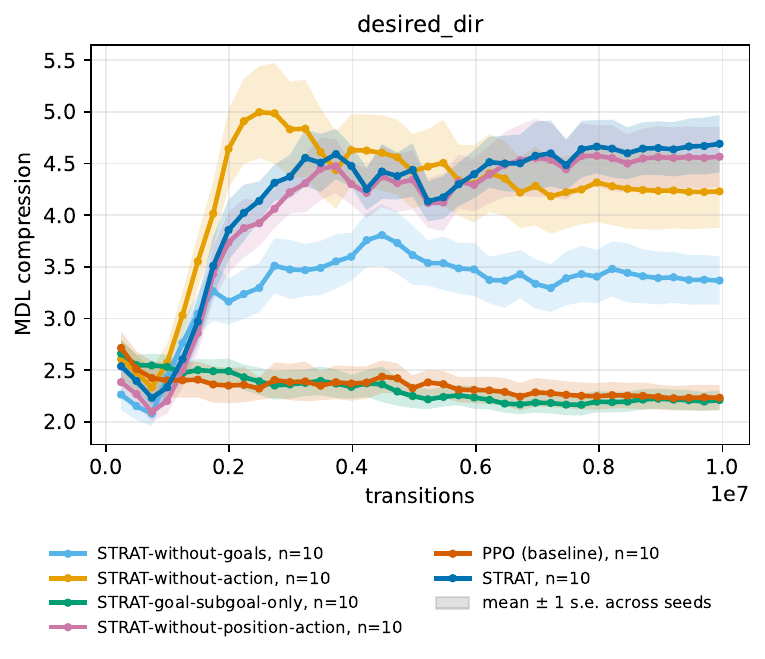}
        \caption{Desired direction compressed in the representation.}
        \label{fig:results_mdl_desired_dir}
    \end{subfigure}\hfill
    \begin{subfigure}[t]{0.48\textwidth}
        \centering
        \includegraphics[width=\linewidth]{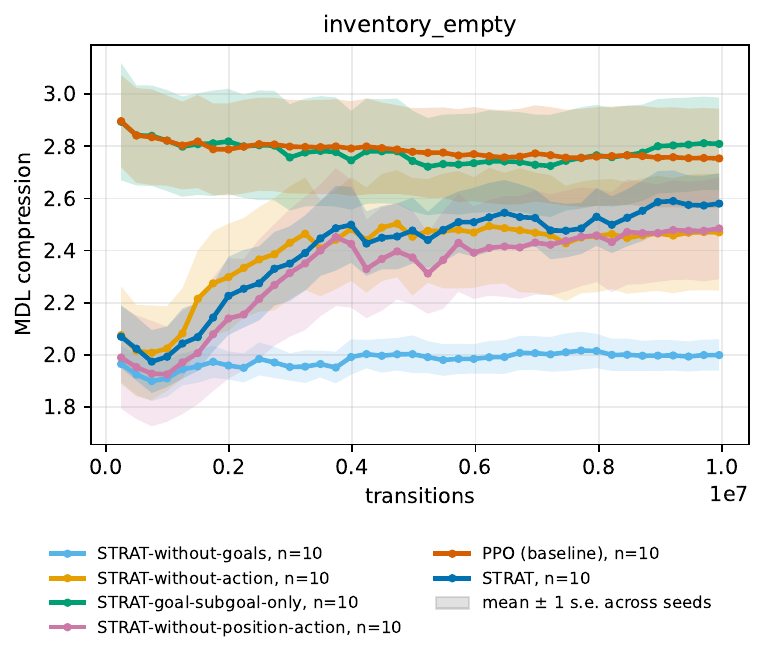}
        \caption{Empty inventory compressed in the representation.}
        \label{fig:results_mdl_inventory_empty}
    \end{subfigure}
    \caption{Minimum description length.}
    \label{fig:results_mdl_1}
\end{figure}

\begin{figure}[t]
    \centering
    \begin{subfigure}[t]{0.48\textwidth}
        \centering
        \includegraphics[width=\linewidth]{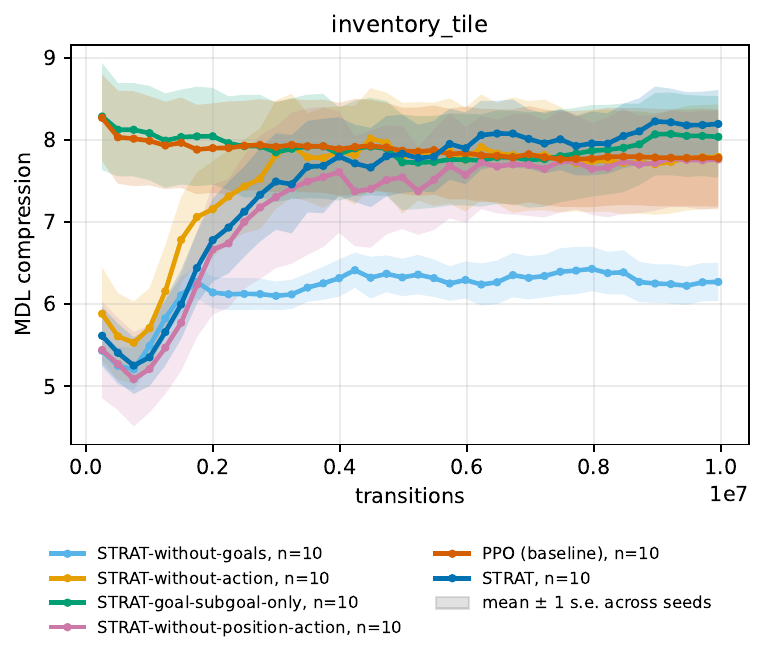}
        \caption{Inventory tile compressed in the representation.}
        \label{fig:results_mdl_inventory_tile}
    \end{subfigure}\hfill
    \begin{subfigure}[t]{0.48\textwidth}
        \centering
        \includegraphics[width=\linewidth]{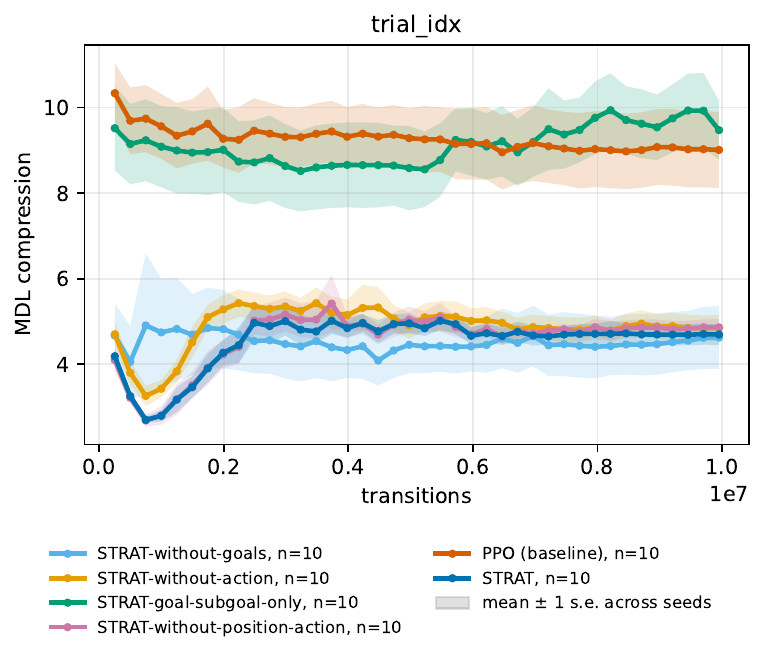}
        \caption{Trial index compressed in the representation.}
        \label{fig:results_mdl_trial_idx}
    \end{subfigure}
    \caption{Minimum description length.}
    \label{fig:results_mdl_2}
\end{figure}

\begin{figure}[t]
    \vspace{0.6em}
    \begin{subfigure}[t]{0.48\textwidth}
        \centering
        \includegraphics[width=\linewidth]{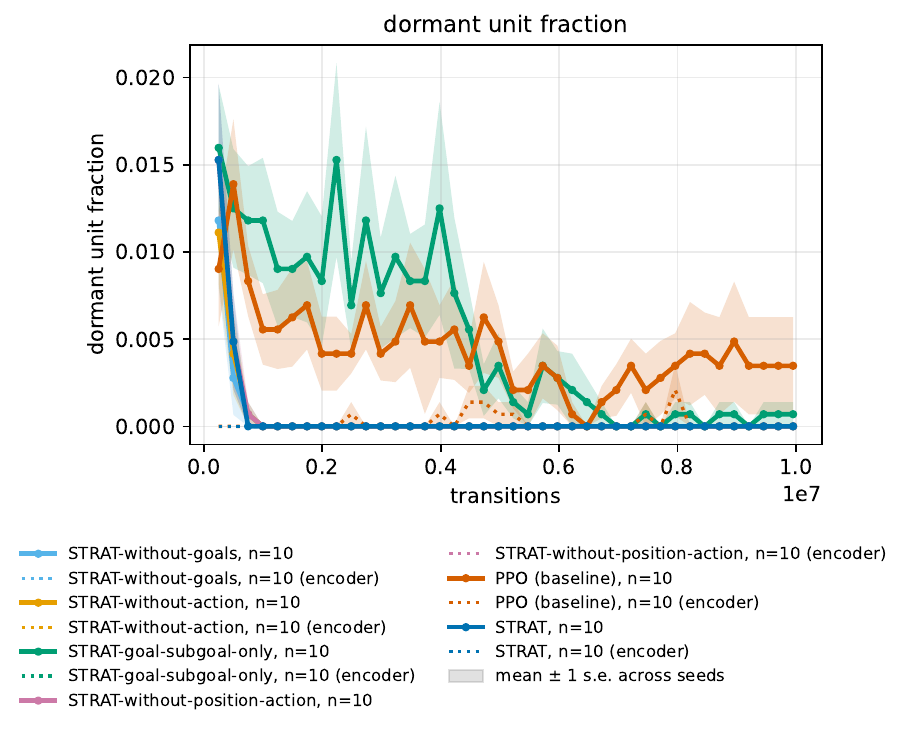}
        \caption{Dormant trunk.}
        \label{fig:results_capacity_dormant}
    \end{subfigure}
    \begin{subfigure}[t]{0.48\textwidth}
        \centering
        \includegraphics[width=\linewidth]{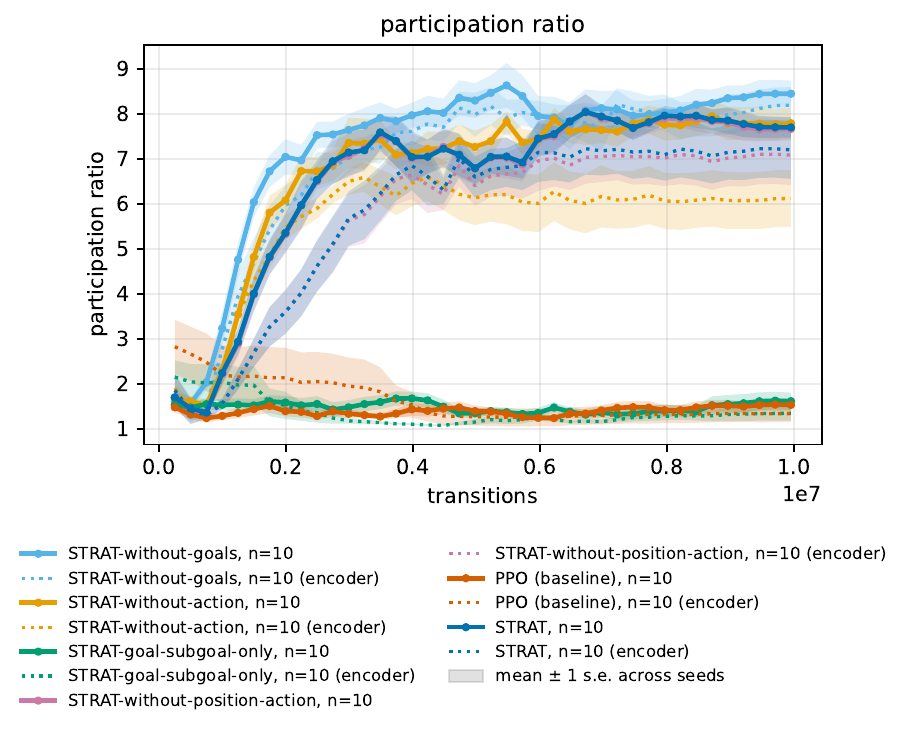}
        \caption{Participation ratio.}
        \label{fig:results_capacity_pr}
    \end{subfigure}
    \caption{Information compressed in the representation capacity.}
    \label{fig:results_capacity}
\end{figure}

\end{document}